\documentclass[runningheads]{llncs}
\usepackage{graphicx}

\usepackage{tikz}
\usepackage{comment}
\usepackage{amsmath,amssymb} 
\usepackage{color}
\usepackage{multirow}
\usepackage{makecell}

\usepackage[accsupp]{axessibility}  

\begin{document}
\pagestyle{headings}
\mainmatter
\def\ECCVSubNumber{4346}  

\title{PillarNet: Real-Time and High-Performance Pillar-based 3D Object Detection} 

\titlerunning{PillarNet}
%
\author{Guangsheng Shi\inst{1} \and
Ruifeng Li\inst{1\star} \and
Chao Ma\inst{2}\thanks{Corresponding authors. Work done while G. Shi visits the Vision and Learning Group at Shanghai Jiao Tong University. }}
\authorrunning{G. Shi, R. Li, and C. Ma}
%
\institute{State Key Laboratory of Robotics and System, Harbin Institute of Technology \\
\and
MoE Key Lab of Artificial Intelligence, AI Institute, Shanghai Jiao Tong University\\
\email{sgsadvance@163.com, lrf100@hit.edu.cn, chaoma@sjtu.edu.cn}}

\maketitle

\begin{abstract}
Real-time and high-performance 3D object detection is of critical importance for autonomous driving. 
Recent top-performing 3D object detectors mainly rely on point-based or 3D voxel-based convolutions, which are both computationally inefficient for onboard deployment.
In contrast, pillar-based methods use solely 2D convolutions, which consume less computation resources, but they lag far behind their voxel-based counterparts in detection accuracy.
In this paper, by examining the primary performance gap between pillar- and voxel-based detectors, we develop a real-time and high-performance pillar-based detector, dubbed PillarNet.  
The proposed PillarNet consists of a powerful encoder network for effective pillar feature learning, a neck network for spatial-semantic feature fusion and the commonly used detect head.
Using only 2D convolutions, PillarNet is flexible to an optional pillar size and compatible with classical 2D CNN backbones, such as VGGNet and ResNet.
Additionally, PillarNet benefits from our designed orientation-decoupled IoU regression loss along with the IoU-aware prediction branch.
Extensive experimental results on the large-scale nuScenes Dataset and Waymo Open Dataset demonstrate that the proposed PillarNet performs well over state-of-the-art 3D detectors in terms of effectiveness and efficiency.
Code is available at \url{https://github.com/agent-sgs/PillarNet}.

\keywords{3D object detection, point cloud, autonomous driving}
\end{abstract}

\section{Introduction}

\begin{figure}[t]
\centering
\includegraphics[]{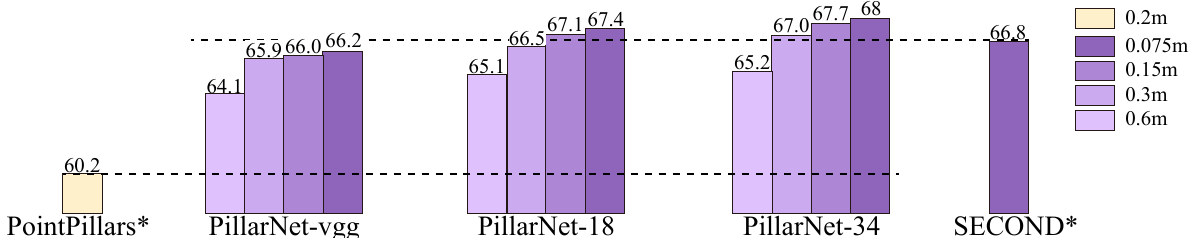}
\caption{Comparison between PillarNet variants along with different pillar sizes and two baselines on the nuScenes \textit{val} set in nuScenes detection score (NDS).
The reported results of these two baselines are from the latest CenterPoint \cite{yin2021center}. All of our PillarNet variants use the same training schedules with CenterPoint-SECOND \cite{yin2021center}.
* denotes the reproduced two baselines using the center-based head from CenterPoint \cite{yin2021center}.
}
\label{fig:performance}
\end{figure}

With the success in point cloud representation learning using deep neural networks, LiDAR-based 3D object detection has made remarkable progress recently.
However, the top-performing point cloud 3D object detectors on the large-scale benchmark datasets, such as nuScenes Dataset \cite{caesar2020nuscenes} and Waymo Open Dataset \cite{sun2020scalability}, entail heavy computational load and large memory storage. Hence, it is desirable to develop a top-performing 3D detector with real-time speed for the onboard deployment on autonomous vehicles. 

Existing point cloud 3D object detectors mainly use the grid-based representation over point cloud and can be broadly categorized into two groups, \textit{i.e.}, 3D voxel-based and 2D pillar-based methods. Both of these two groups take the classical ``encoder-neck-head" detection architecture \cite{he2020structure,Hu2021AFDetV2RT,lang2019pointpillars,noh2021hvpr,wang2019voxel,yan2018second,ye2020hvnet,zheng2020ciassd,zhou2018voxelnet}.
Voxel-based methods \cite{he2020structure,Hu2021AFDetV2RT,yan2018second,zheng2020ciassd,zhou2018voxelnet} typically divide the input point cloud into regular 3D voxel grid. 
An encoder with sparse 3D convolutions \cite{graham2017submanifold} is then used to learn geometric representation across multiple levels. Following the encoder, a neck module with standard 2D CNNs fuses multi-scale features before feeding to the detection head.
In contrast, pillar-based methods \cite{lang2019pointpillars,wang2019voxel,ye2020hvnet,noh2021hvpr} project 3D point clouds into a 2D pseudo-image on the BEV plane, and then directly build the neck network upon the 2D CNN-based feature pyramid network (FPN) to fuse multi-scale features.
For voxel-based methods, the effective voxel-wise feature learning powered by sparse 3D CNN delivers favorable detection performance.
However, due to the 3D sparse convolution within the encoder, it is hard to aggregate multi-scale features with different resolutions on the BEV space.
For pillar-based methods, a light encoder for pillar feature learning yields unsatisfied performance compared with their voxel-based counterparts.
Moreover, the small sized pseudo-image and the large initial pillar further limit the detection performance.
It is because a finer pillar leads to larger pseudo-image and more favorable performance but heavier computational load.
Interestingly, both voxel- and pillar-based methods perform 3D detection using the aggregated multi-scale features on the BEV space (see Sec. \ref{sec:preliminary}) 

We observe that previous pillar-based methods do not have powerful pillar feature encoding, which is the main cause of the unsatisfied performance. In addition, progressively downsampling pillar scales can help to decouple the output feature map size and the initial pseudo-image projection scale.
As such, we design a real-time and high-performance pillar-based 3D detection method, dubbed PillarNet, that consists of an encoder for hierarchical deep pillar feature extraction, a neck module for multi-scale feature fusion, and the commonly-used center-based detect head.
In our PillarNet, the powerful encoder network involves 5 stages. Stage 1 to 4 follow the same setting as the conventional 2D detection networks such as VGG \cite{simonyan2014very} and ResNet \cite{he2016deep} but substituted 2D convolutions with its sparse counterparts for resource savings. Stage 5 with standard 2D convolutions possesses a larger receptive field and feeds semantic features to the following neck network.
The neck network exchanges sufficient information through stacked convolution layers between the further enriched high-level semantic feature from the encoder stage 5 and the low-level spatial feature from the encoder stage 4.
For tuning the hard-balanced pillar size in previous pillar-based methods, PillarNet offers an effective solution by skillfully detaching the corresponding encoder stages for the chosen pillar scale. For example, to accommodate the input with 8 times pillar size ($0.075*8$m in nuScenes Dataset), we can simply remove the 1x, 2x, and 4x downsampled encoder stages.

As shown in Fig. \ref{fig:performance}, our PillarNet with variant configurations, \textit{i.e.}, PillarNet-vgg/18/34, offer the scalability and flexibility for point cloud-based 3D object detection by using merely 2D convolutions.
Our PillarNet significantly advances pillar-based 3D detectors and sheds new light on further research on point cloud object detection.
Despite its simplicity, the proposed PillarNet achieves the state-of-the-art performance on two large-scale autonomous driving benchmarks \cite{caesar2020nuscenes,sun2020scalability} and runs in real-time (see Sec. \ref{sec:experiment}).

\section{Related Works}

\subsection{Point Cloud 3D Object Detection}

3D object detection with point cloud alone can mainly be summarized into two categories: point-based and grid-based methods.

{\flushleft \bf Point-based 3D object detectors.}
Powered by the pioneering PointNet \cite{qi2017pointnet,qi2017pointnet++}, point-based methods directly process irregular point clouds and predict 3D bounding boxes.
PointRCNN \cite{shi2019pointrcnn} proposes a point-based proposal generation paradigm directly from raw point clouds and then refines each proposal by devising an RoI pooling operation. 
STD \cite{yang2019std} transforms point features inside of each proposal into compact voxel representation for RoI feature extraction.
3DSSD \cite{liu2016ssd}, as a one-stage 3D object detector, introduces F-FPS as a complement of existing D-FPS with set abstraction operation to benefit both regression and classification.
These point-based methods naturally preserve accurate point location and enable flexible receptive fields with radius-based local feature aggregation. These methods, however, as summarized in \cite{liu2019point}, spend 90\% of their runtime on organizing irregular point data rather than extracting features, and are not suitable for handling large-scale point clouds.

{\flushleft \bf Grid-based 3D object detectors.}
Most existing methods discrete the sparse and irregular point clouds into regular grids including 3D voxels and 2D pillars, and then capitalize on 2D/3D CNN to perform 3D object detection.
The pioneering VoxelNet \cite{zhou2018voxelnet} divides point cloud into 3D voxels, and encodes scene feature using 3D convolutions.
To tackle the empty voxels typically for the large outdoor space, SECOND \cite{yan2018second} introduces 3D sparse convolution to accelerate VoxelNet \cite{zhou2018voxelnet} and improves the detection accuracy. 
Until now, 3D voxel-based methods dominate the majority of 3D detection benchmarks. 
For a long time, even with sparse 3D convolution, it was hard to balance between the fine resolution of 3D voxels and associated resource costs.

PointPillars \cite{lang2019pointpillars} uses 2D voxelization on the ground plane with a PointNet \cite{qi2017pointnet} based per-pillar feature extractor. It can utilize 2D convolutions for deployment on embedded systems with limited costs.
MVF \cite{chen2017multi} utilizes multi-view features to augment point-wise information before projecting raw points into 2D pseudo-image. 
HVNet \cite{ye2020hvnet} fuses different scales of pillar features at the point-wise level to achieve good accuracy and high inference speed.
HVPR\cite{noh2021hvpr} cleverly keeps the efficiency of pillar-based detection while implicitly leveraging the voxel-based feature learning regime  for better performance.
Current Pillar-based advancements, however, focus on the sophisticated pillar feature projection or multi-scale aggregation strategies, to narrow the huge performance gap relative to their voxel-based counterparts.
In contrast, we resort to a powerful backbone network to address the above issues and boost 3D detection performance.

\subsection{Multi-sensor based 3D Object Detection}

Most approaches expect the complementary information from multiple sensors, such as camera image and LiDAR, to achieve high performance 3D object detection.
MV3D \cite{chen2017multi} designs 3D object anchors and generates proposals from BEV representations and refine them using features from LiDAR and camera. 
AVOD \cite{ku2018joint} instead fuses these features at the proposal generation stage and provides better detection results.
ContFuse \cite{liang2018deep} learns to fuse image features with point cloud features onto BEV space. 
MMF \cite{liang2019multi} struggles for LiDAR-Camera feature fusion via proxy tasks including depth completion on RGB images and ground estimation from point clouds. 
3D-CVF \cite{yoo20203d} tackles the multi-sensor registration issue for cross-view spatial feature fusion in the BEV domain.
Almost all of these multi-modality frameworks rely on the intermediate BEV representation to perform 3D object detection.
Our method extracts point cloud features on BEV space and may be promising to seamlessly integrate into existing multi-modality frameworks for advanced performance.

\section{PillarNet for 3D Object Detection}

\subsection{Preliminaries}
\label{sec:preliminary}

\begin{figure}[!t]
\centering
\setlength{\tabcolsep}{1mm}
\begin{tabular}{ccc}
\includegraphics[]{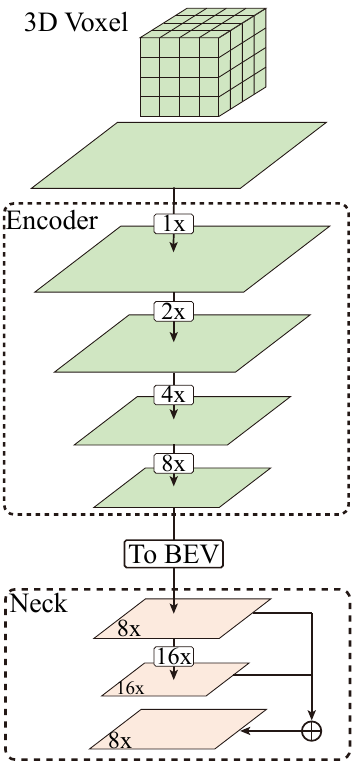} & 
\includegraphics[]{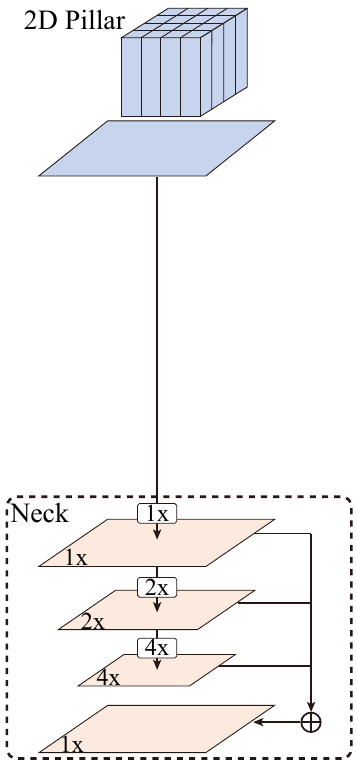} & 
\includegraphics[]{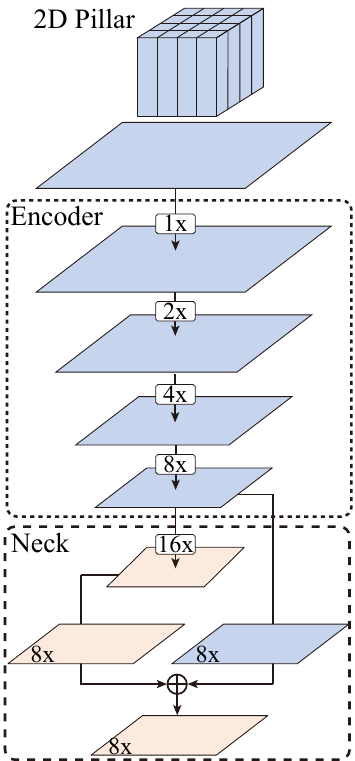} \\
A: SECOND & B: PointPillars & C: PillarNet (Ours) \\
\end{tabular}
\caption{Comparison of three types of architectures. The encoder uses sparse 3D CNN in SECOND \cite{yan2018second} while sparse 2D CNN for PillarNet. The neck in all the three methods uses standard 2D CNN.
On the nuScenes Dataset, the 3D voxel size in SECOND \cite{yan2018second} is (0.075m, 0.075m, 0.2m), and the 2D pillar size in PointPillars and our proposed PillarNet is (0.2m, 0.2m) and (0.075m, 0.075m) respectively.}
\label{fig:overview}
\end{figure}

The grid-based detectors perform 3D detection on BEV space, including 3D voxel-based detectors and 2D pillar-based detectors.
Recent voxel-based detectors follow the SECOND \cite{yan2018second} architecture with improved sparse 3D CNN for effective voxel feature encoding over the pioneering VoxelNet \cite{zhou2018voxelnet}.
Pillar-based detectors generally follow the pioneering PointPillars \cite{lang2019pointpillars} architecture with only 2D CNN for multi-scale feature fusion.
We first revisit these two representative point cloud detection architectures, which motivate us to construct the proposed PillarNet method.

{\flushleft \bf SECOND.} SECOND \cite{yan2018second} is a typical voxel-based one-stage object detector, which lays the groundwork for succeeding voxel-based detectors with specialized sparse 3D convolutions \cite{graham20183d,graham2017submanifold}.
It divides the unordered point cloud into regular 3D voxels and performs box prediction on BEV space.
The entire 3D detection architecture contains three basic parts: 
(1) An encoder hierarchically encodes the input non-empty voxel features into 3D feature volumes with the $1\times$, $2\times$, $4\times$ and $8\times$ downsampled sizes.
(2) A neck module further abstracts the flattened encoder output on the BEV space into multiple scales in a top-down manner.
(3) A detect head performs box classification and regression using the fused multi-scale BEV features.

{\flushleft \bf PointPillars.}
PointPillars \cite{lang2019pointpillars} projects raw point cloud on the X-Y plane via a tiny PointNet \cite{qi2017pointnet}, yielding a sparse 2D pseudo-image. 
PointPillars uses a 2D CNN-based top-down network to process the pseudo-image with stride 1x, 2x, and 4x convolution blocks and then  concatenates the multi-scale features for the detect head.

{\flushleft \bf Analysis.}
Despite the favorable runtime and memory efficiency, PointPillar  \cite{lang2019pointpillars} still lags far behind SECOND \cite{yan2018second} on performance.
Under the premise that sparse 3D convolutions possess the superior representation ability for point cloud learning, recent advanced pillar-based methods mainly focus on exploring attentive pillar feature extraction \cite{liu2020tanet,ye2020hvnet} from raw points or sophisticated multi-scale strategies \cite{wang2019voxel,ye2020hvnet,noh2021hvpr}. These methods, on the other hand, suffer from unfavourable latency and still under-performs their 3D voxel-based counterparts by a large margin.

Alternatively, we take a different view by considering grid-based detectors as BEV-based detectors and revisit the entire point cloud learning architecture.  
We identify that the performance bottleneck of pillar-based methods mainly lies in the sparse encoder network for spatial feature learning and effective neck module for sufficient spatial-semantic features fusion.
Specifically, PointPillars directly applies the feature pyramid network to fuse multi-scale features on the projected dense 2D pseudo-image, lacking the sparse encoder network for effective pillar feature encoding as in SECOND.
On the other hand, PointPillars couples the size of the final output feature maps with the initial projected pillar scale, increasing the entire calculation and memory cost sharply as the pillar scale gets finer.

To resolve the above issues, we stand by the ``encoder-neck-head" detection architecture on BEV space to improve the performance of pillar-based methods.
Specifically, we explore the significant difference and respective function for the encoder and neck networks:
\begin{enumerate}
\item[-] We redesign the encoder in SECOND by substituting sparse 3D convolutions by its sparse 2D convolutions counterpart on loss-less pillar features from raw point clouds. It has been validated in the $3^{rd}$ row of Table \ref{tab:ab_compare} that the sparse encoder process enhances 3D detection performance significantly.
\item[-] We formulate the neck module as the spatial-semantic feature fusion by inheriting the sparse spatial features from the sparse encoder output and further high-level semantic feature abstraction in low-resolution feature maps, as shown in the $6^{th}$ row of Table \ref{tab:ab_compare}, which is efficient and effective.
\end{enumerate}
Finally, we build our PillarNet using the relatively heavyweight sparse encoder network for hierarchical pillar feature learning and the lightweight neck module for sufficient spatial-semantic feature fusion.

\subsection{PillarNet Design for 3D Object Detection}

\begin{figure}
\centering
\includegraphics[width=\textwidth]{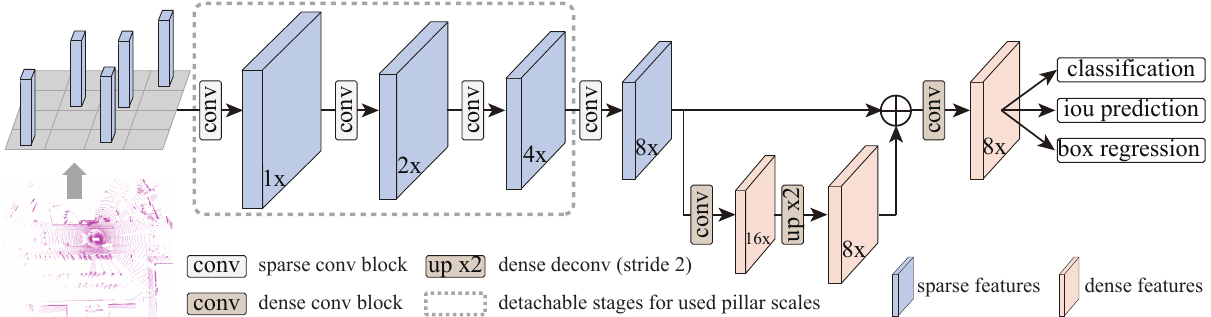}
\caption{The overall architecture of our proposed PillarNet. The input point clouds are first quantified into pillars to feed into the 2D sparse convolution-based encoder to learn multi-scale spatial features. Then the densified semantic feature is fused with the spatial feature in the neck module for the final 3D box regression, classification, and IoU prediction.}
\label{fig:pipeline}
\end{figure}

In this subsection, we present the detailed structure of our PillarNet design.
The overall architecture in Fig.~\ref{fig:pipeline} consists of three components: the encoder for deep pillar feature extraction, the neck module for spatial-semantic feature aggregation, and the 3D detect head. 
With the commonly used center-based detect head \cite{yin2021center}, we  present the flexibility and scalability of our PillarNet.

{\flushleft \bf Encoder design.} 
The encoder network aims to extract deep sparse pillar features hierarchically from the projected sparse 2D pillar features, where the detachable stages from 1 to 4 progressively down-sample sparse pillar features using sparse 2D CNN.
Compared to PointPillars\cite{lang2019pointpillars}, our designed encoder have two advantages:
\begin{itemize}\itemsep2pt
\item[(1)] The sparse encoder network can take the progress on image-based 2D object detection, such as VGGNet \cite{simonyan2014two} and ResNet \cite{he2016deep}.
The simple encoder for pillar feature learning can largely improve 3D detection performance.
\item[(2)] The hierarchically downsampling structure allows PillarNet to skillfully operate the sparse pillar features with different pillar sizes, which alleviates the limitation of coupling pillar size in previous pillar-based methods.
\end{itemize}

Our constructed PillarNet with variant backbones, PilllarNet-vgg/18/34, with the similar complexities of VGGNet/ResNet-18/ResNet-34. The detailed network configurations can be found in the supplementary material.

\begin{figure}[tpb]
\centering
\begin{tabular}{ccc}
\includegraphics[]{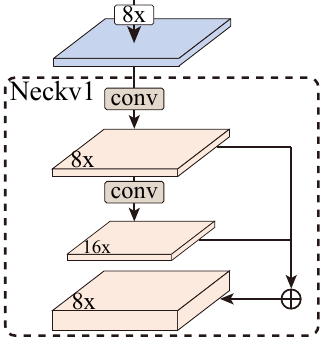} & 
\includegraphics[]{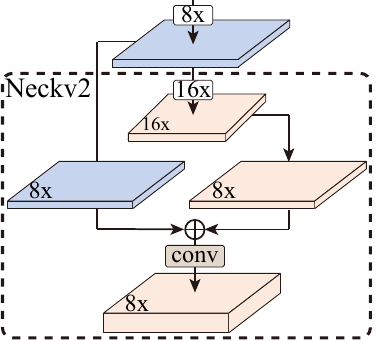} & 
\includegraphics[]{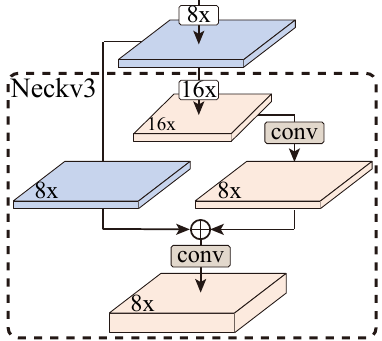} \\
A: A naive neck design 
& \makecell{B: An aggressive neck design \\ using the 16x features}
& \makecell{C: A enriched neck design \\ using the 16x features}  \\
\end{tabular}
\caption{Detailed structure of different neck designs. 
The neck A inherits directly from SECOND \cite{yan2018second}, while two alternative neck designs B and C introduce the spatial features from sparse encoder network and semantic features from the 16$\times$  dense feature maps.}
\label{fig:neck}
\end{figure}

{\flushleft \bf Neck design.}
The neck module, as in FPN \cite{lin2017feature}, aims to fuse high-level abstract semantic features and low-level fine-grained spatial features for mainstream detect head (\textit{i.e.}, anchor boxes or anchor points).
The additional 16X downsampled dense feature maps further abstracts high-level semantic feature using a group of dense 2D CNNs, to enrich receptive field for large objects and populate object center-positioned features for center-based detect head.
Equipped with spatial features from sparse encoder network, there are two alternative neck designs for the spatial-semantic feature fusion from the starting design in SECOND \cite{yan2018second}:
\begin{itemize}\itemsep2pt
\item[(1)] The naive design neckv1 (Fig. \ref{fig:neck}(A)) from SECOND \cite{yan2018second} applies a top-down network to generate multi-scale features and concatenate multi-scale dense feature maps as the final output.
\item[(2)] The aggressive design neckv2 (Fig. \ref{fig:neck}(B)) considers sufficient information exchange between high-level semantic feature from additional 16X downsampled dense feature maps and low-level spatial feature from sparse encoder network using a group of convolution layers.
\item[(3)] The design neckv3 (Fig. \ref{fig:neck}(C)) further enriches the high-level semantic features on 16X downsampled dense feature maps through a group of convolution layers and fuses the spatial-semantic features with the other group of convolution layers for robust feature extraction.
\end{itemize}

\subsection{Orientation-Decoupled IoU Regression Loss}
In general, the IoU metric highly correlates with the localization quality and classification accuracy of the predicted 3D boxes. Previous methods \cite{liang2021rangeioudet}
show that using the 3D IoU quality to re-weight the classification and supervise the box regression can achieve better localization accuracy. 

For the classification branch, we follow previous methods \cite{Hu2021AFDetV2RT,zheng2020ciassd} and use the IoU-rectification scheme to incorporate the IoU information into the confidence scores.
The IoU-Aware rectification function \cite{Hu2021AFDetV2RT} at the post-processing stage can be formulated as:
\begin{equation}
\centering
\hat{S} = S^{1-\beta} * W_{\rm{IoU}}^{\beta}
\end{equation}
where $S$ indicates the classification score and $W_{\rm{IoU}}$ is the IoU score. $\beta$ is a hyper-parameter.
For predicting the IoU score, we use L1 loss $\mathcal{L}_{iou}$ to supervise the IoU regression, where the target 3D IoU score $W$ between the predicted 3D box and the ground truth box is encoded by $2 * (W - 0.5) \in [-1, 1]$.

\begin{figure}[t]
\centering
\begin{tabular}{cccc}
\includegraphics[width=0.22\textwidth]{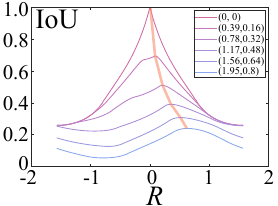} & 
\includegraphics[]{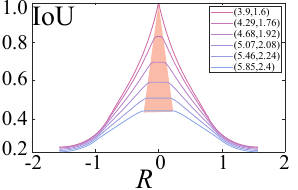} & 
\includegraphics[width=0.22\textwidth]{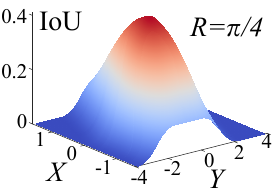} & 
\includegraphics[width=0.28\textwidth]{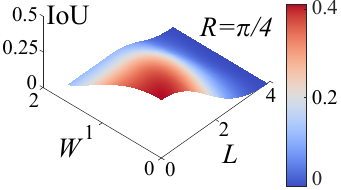} \\
\makecell{A: IoU-\textit{R} curve for \\ center deviation} & 
\makecell{B: IoU-\textit{R} curve for \\ size variation} &
\makecell{C: IoU-XY surface \\ when $R=\pi/4$} & 
\makecell{D: IoU-WL surface \\ when $R=\pi/4$} \\
\end{tabular}
\caption{The IoU metric-based interplay of orientation (\textit{R}) with center or size for a 2D rotated box [0, 0, 3.9, 1.6, 0]. A and B depict the effect of center variation and size oscillation on orientation regression separately. Red curve in A indicates the local optimum while red region in B for optimization plateau.
C and D depict the effect of orientation bias $R=\pi/4$ on center and size regression separately.}
\label{fig:orientation}
\end{figure}

For the regression branch, 
recent methods \cite{liang2021rangeioudet,zheng2021se} extend the GIoU \cite{rezatofighi2019generalized} loss or DIoU \cite{zheng2020distance} loss from 2D detection to the 3D domain. 
However, the non-trivial 3D IoU computation slows down the training process. Furthermore, the coupled orientation for the IoU-related regression may negatively affect the training process.
Fig. \ref{fig:orientation} shows such an example. Given a typical 2D bounding box $[x,y,l,w,\theta]=[0, 0, 3.9, 1.6, 0]$, there exist cross-effects of orientation with center bias for $x$ and $y$ positions or of scale for width and length sizes during the optimization for IoU metric between biased box with ground-truth box as follows:
\begin{itemize}
\item[-] The effect of center deviation on orientation regression.  
The training phase easily settles into a local optimum, if the box center deviates far. See the red curve in Fig. \ref{fig:orientation}(A).
\item[-] The effect of size variation on orientation regression. 
The training phase settles into the notorious optimization plateau, if box sizes change largely. See the red region in Fig. \ref{fig:orientation}(B).
\item[-] The effect of orientation bias on center and size regression. The optimization direction remains consistent even if the orientation deviation is significant.
\end{itemize}

As a result, we present an alternative Orientation-Decoupled IoU-related regression loss by decoupling the orientation $\theta$ from the mutually-coupled seven parameters $(x, y, z, w, l, h, \theta)$.
Specifically, we extend the IoU regression loss $\mathcal{L}_{od-iou}$ (OD-IoU/OD-GIoU/OD-DIoU) from the IoU loss \cite{rezatofighi2019generalized}, GIoU loss \cite{rezatofighi2019generalized} and DIoU loss \cite{zheng2020distance}, respectively.

\subsection{Overall Loss Function}

Following \cite{yin2021center}, we apply the focal loss \cite{lin2017focal} for the heatmap classification $\mathcal{L}_{cls}$, and the L1 loss for localization offset $\mathcal{L}_{off}$, the z-axis location $\mathcal{L}_{z}$, 3D object size $\mathcal{L}_{size}$ and orientation $\mathcal{L}_{ori}$.
The overall loss $\mathcal{L}_{total}$ is jointly optimized as follows:
\begin{equation}
\centering
\mathcal{L}_{total} = \mathcal{L}_{cls} + \mathcal{L}_{iou} + \lambda (\mathcal{L}_{od-iou} + \mathcal{L}_{off} + \mathcal{L}_{z} + \mathcal{L}_{size} + \mathcal{L}_{ori})
\label{eq:one-loss}
\end{equation}
where the loss weight $\lambda$ is empirically set parameter as in \cite{yin2021center}.

\section{Experiments}
\label{sec:experiment}

{\flushleft \bf nuScenes Dataset.}
nuScenes \cite{caesar2020nuscenes} contains 1000 driving sequences, with 700, 150, 150 sequences for training, validation, and testing, respectively. Each sequence is approximately 20-second long, with a LiDAR frequency of 20 FPS. 
nuScenes uses a 32 lanes LiDAR, which produces approximately 30k points per frame. The annotations include 10 classes with a long-tail distribution. The official evaluation metrics are mean Average Precision (mAP) and nuScenes detection score (NDS). 
We follow the convention to accumulate 10 LiDAR sweeps to densify the point clouds and report results by using the official evaluation protocol.

{\flushleft \bf Waymo Open Dataset.}
Waymo Open Dataset \cite{sun2020scalability} is currently the largest dataset with LiDAR point clouds for autonomous driving. There are total 798 training sequences with around 160k LiDAR samples, and 202 validation sequences with 40k LiDAR samples. It annotated the objects in the full 360$^{\circ}$ field. The evaluation metrics are calculated by the official evaluation tools, where the mean average precision (mAP) and the mean average precision weighted by heading (mAPH) are used for evaluation. The 3D IoU threshold is set as 0.7 for vehicle detection and 0.5 for pedestrian/cyclist detection. 

{\flushleft \bf Training and Inference details.}
We use the same training schedules as prior CenterPoint-SECOND \cite{yin2021center}, where Adam optimizer is used with one-cycle learning rate policy, weight decay 0.01, and momentum 0.85 to 0.95 on 4 Tesla V100 GPUs.
We make runtime comparison with two baselines (\textit{i.e., CenterPoint-SECOND and CenterPoint-PointPillars}) on desktop equipped with an i9 CPU and RTX 3090 GPU.
To project raw point clouds into the pillar feature, we apply one-layer MLP-based PointNet associated with $atomic \ max$-based pooling on augmented point-wise feature of all inside points per pillar.
We adopt the widely used data augmentation strategies as \cite{yin2021center} during training, including the random scene flipping along, random rotation, random scene scaling, and random translation.


For nuScenes Dataset, 
we set the detection range to $[-54m, 54m]$ for the X and Y axis, and $[-5m, 3m]$ for the Z axis. 
We use $(0.075m, 0.075m)$ as the basic pillar size for experiments.
We train the PillarNet from scratch with batch size 16, max learning rate 1e-3 for 20 epochs.
For the post-processing process during inference, following \cite{yin2021center}, we use class-agnostic NMS with the score threshold set to 0.1 and rectification factor $\beta$ to 0.5 for all 10 classes.
To compare on the nuScenes test set, we do not use any model ensembling except double-flip test-time augmentation as CenterPoint \cite{yin2021center}.

For Waymo Open Dataset, 
we set the detection range to $[-75.2m, 75.2m]$ for X and Y axes, and $[-2m, 4m]$ for Z axis.
W use $(0.1m, 0.1m)$ as the basic pillar size for experiments and train the PillarNet from scratch with batch size 16, max learning rate 3e-3 for 36 epochs.
During inference, we simply follow \cite{Hu2021AFDetV2RT} by using class-specific NMS with the IoU thresholds (0.8, 0.55, 0.55) and rectification factor $\beta$ to (0.68, 0.71, 0.65) for vehicle, pedestrian and cyclist respectively.

\setlength{\tabcolsep}{1pt}
\begin{table}[t]
\centering
\caption{The LiDAR-only non-ensemble 3D detection performance comparison on the nuScenes \textit{test} set. The table is mainly sorted by nuScenes detection score (NDS) which is the official ranking metric.}
\label{tab:nuscenes_test}
\resizebox{\textwidth}{!}{%
\begin{tabular}{c|c|cc|c|c|c|c|c|c|c|c|c|c}
\hline\noalign{\smallskip}
Methods & Stages & NDS & mAP & Car & Truck & Bus & Trailer & Cons.Veh. & Ped. & Motor. & Bicycle & Tr.Cone & Barrier  \\ 
\noalign{\smallskip}
\hline
\noalign{\smallskip}
WYSIWYG \cite{hu2020you} & One & 41.9 & 35.0 & 79.1 & 30.4 & 46.6 & 40.1 & 7.1 & 65.0 & 18.2 & 0.1 & 28.8 & 34.7 \\
PointPillars \cite{lang2019pointpillars} & One & 45.3 & 30.5 & 68.4 & 23.0 & 28.2 & 23.4 & 4.1 & 59.7 & 27.4 & 1.1 & 30.8 & 38.9 \\ 
3DVID \cite{yin2020lidar} & One & 53.1 & 45.4 & 79.7 & 33.6 & 47.1 & 43.1 & 18.1 & 76.5 & 40.7 & 7.9 & 58.8 & 48.8 \\
3DSSD \cite{yang20203dssd} & One & 56.4 & 42.6 & 81.2 & 47.2 & 61.4 & 30.5 & 12.6 & 70.2 & 36.0 & 8.6 & 31.1 & 47.9 \\
Cylinder3D \cite{zhu2021cylindrical} & One & 61.6 & 50.6 & - & - & - & - & - & - & - & - & - & - \\
CGBS \cite{zhu2019class} & One & 63.3 & 52.8 & 81.1 & 48.5 & 54.9 & 42.9 & 10.5 & 80.1 & 51.5 & 22.3 & 70.9 & 65.7 \\
CVCNet \cite{chen2020every} & One & 64.2 & 55.8 & 82.6 & 49.5 & 59.4 & 51.1 & 16.2 & 83.0 & 61.8 & 38.8 & 69.7 & 69.7 \\
CenterPoint \cite{yin2021center} & Two & 65.5 & 58.0 & 84.6 & 51.0 & 60.2 & 53.2 & 17.5 & 83.4 & 53.7 & 28.7 & 76.7 & 70.9 \\
HotSpotNet \cite{chen2020object} & One & 66.0 & 59.3 & 83.1 & 50.9 & 56.4 & 53.3 & 23.0 & 81.3 & 63.5 & 36.6 & 73.0 & 71.6 \\
AFDetV2 \cite{Hu2021AFDetV2RT} & One & 68.5 & 62.4 & 86.3 & 54.2 & 62.5 & 58.9 & 26.7 & 85.8 & 63.8 & 34.3 & 80.1 & 71.0 \\
\hline
PillarNet-vgg & One & 69.6 & 63.3 & 86.9 & 56.0 & 62.2 & 62.0 & 28.6 & 86.3 & 62.6 & 33.5 & 79.6 & 75.6 \\
PillarNet-18 & One & 70.8 & 65.0 & 87.4 & 56.7 & 60.9 & 61.8 & \textbf{30.4} & 87.2 & 67.4 & 40.3 & 82.1 & 76.0 \\
PillarNet-34 & One & \textbf{71.4} & \textbf{66.0} & \textbf{87.6} & \textbf{57.5} & \textbf{63.6} & \textbf{63.1} & {27.9} & \textbf{87.3} & \textbf{70.1} & \textbf{42.3} & \textbf{83.3} & \textbf{77.2} \\
\hline 
\end{tabular}
}
\end{table}

\begin{table}[thb]
\centering
\caption{Single- (upper group) and multi-frame (lower group) LiDAR-only non-ensemble performance comparison on the Waymo Open Dataset \textit{test} set. 
"L" and "LT" mean "all LiDARs" and "top-LiDAR only", respectively.
$\dagger$ denotes the reported results from RSN \cite{sun2021rsn}.
}
\label{tab:waymo_test}
\resizebox{\textwidth}{!}{%
\begin{tabular}{c|c|c|c|cc|cc|cc|cc|cc|cc}
\hline\noalign{\smallskip}
\multirow{2}{*}{Methods} & \multirow{2}{*}{Stages} & \multirow{2}{*}{Sensors} & \multirow{2}{*}{Frames} & \multicolumn{2}{c|}{Vehicle (L1)} & \multicolumn{2}{c|}{Vehicle (L2)} & 
\multicolumn{2}{c|}{Ped. (L1)} & \multicolumn{2}{c|}{Ped. (L2)} & 
\multicolumn{2}{c|}{Cyc. (L1)} & \multicolumn{2}{c}{Cyc. (L2)} \\
 & & & & mAP & mAPH & mAP & mAPH & mAP & mAPH & mAP & mAPH & mAP & mAPH & mAP & mAPH \\ 
\noalign{\smallskip}
\hline
\noalign{\smallskip}
$\dagger$ PointPillars \cite{lang2019pointpillars} & One & LT & 1 & 68.60 & 68.10 & 60.50 & 60.10 & 68.00 & 55.50 & 61.40 & 50.10 & - & - & - & - \\
RCD \cite{bewley2020range} & Two & - & 1 & 71.97 & 71.59 & 65.06 & 64.70 & - & - & - & - & - & - & - & - \\
CenterPoint \cite{yin2021center} & Two & LT & 1 & 80.20 & 79.70 & 72.20 & 71.80 & 78.30 & 72.10 & 72.20 & 66.40 & - & - & - & - \\
AFDetV2 \cite{Hu2021AFDetV2RT} & One & LT & 1 & 80.49 & 80.43 & 72.98 & 72.55 & 79.76 & \textbf{74.35} & 73.71 & \textbf{68.61} & \textbf{72.43} & \textbf{71.23} & \textbf{69.84} & \textbf{68.67} \\
PillarNet-vgg & One & LT & 1& 81.16 & 80.68 & 73.64 & 73.20 & 78.30 & 70.28 & 72.23 & 64.68 & 67.26 & 66.07 & 64.79 & 63.65 \\
PillarNet-18 & One & LT & 1 & 81.85 & 81.40 & 74.46 & 74.03 & 79.97 & 72.68 & 73.95 & 67.09 & 67.98 & 66.80 & 65.50 & 64.36 \\
PillarNet-34 & One & LT & 1& \textbf{82.47} & \textbf{82.03} & \textbf{75.07} & \textbf{74.65} & \textbf{80.82} & 74.13 & \textbf{74.83} & \textbf{68.54} & 69.08 & 67.91 & 66.60 & 65.47 \\\hline\hline
3D-MAN \cite{yang20213d} & Multi & L & 15 & 78.71 & 78.28 & 70.37 & 69.98 & 69.97 & 65.98 & 63.98 & 60.26 & - & - & - & - \\
RSN \cite{sun2021rsn} & Two & LT & 3 & 80.70 & 80.30 & 71.90 & 71.60 & 78.90 & 75.60 & 70.70 & 67.80 & - & - & - & - \\
CenterPoint \cite{yin2021center} & Two & L & 2 & 81.05 & 80.59 & 73.42 & 72.99 & 80.47 & 77.28 & 74.56 & 71.52 & 74.60 & 73.68 & 72.17 & 71.28 \\
Pyramid R-CNN \cite{mao2021pyramid} & Two & L & 2 & 81.77 & 81.32 & 74.87 & 74.43 & - & - & - & - & - & - & - & - \\
AFDetV2 \cite{Hu2021AFDetV2RT} & One & LT & 2 & 81.65 & 81.22 & 74.30 & 73.89 & 81.26 & 78.05 & 75.47 & 72.41 & \textbf{76.41} & \textbf{75.37} & \textbf{74.05} & \textbf{73.04} \\
PillarNet-vgg & One &LT & 2 & 82.18 & 81.73 & 74.93 & 74.49 & 80.41 & 76.86 & 74.52 & 71.14 & 68.75 & 67.89 & 66.52 & 65.68 \\
PillarNet-18 & One &LT & 2 & 82.68 & 82.25 & 75.53 & 75.12 & 81.71 & 78.29 & 75.91 & \textbf{72.66} & 70.19 & 69.30 & 68.01 & 67.15 \\
PillarNet-34 & One & LT & 2 & \textbf{83.23} & \textbf{82.80} & \textbf{76.09} & \textbf{75.69} & \textbf{82.38} & \textbf{79.02} & \textbf{76.66} & \textbf{73.46} & 71.44 & 70.51 & 69.20 & 68.29  \\
\hline       
\end{tabular}
}
\end{table}

\begin{table}[th]
\centering
\caption{The single-frame LiDAR-only non-ensemble 3D AP/APH performance comparison on the Waymo Open Dataset \textit{val} set.
$\dagger$: reported by \cite{li2021lidar}.}
\label{tab:waymo_val}
\resizebox{\textwidth}{!}{%
\begin{tabular}{c|c|cc|cc|cc|cc|cc|cc}
\hline\noalign{\smallskip}
\multirow{2}{*}{Methods} & \multirow{2}{*}{Stages} & \multicolumn{2}{c|}{Vehicle (L1)} & \multicolumn{2}{c|}{Vehicle (L2)} & 
\multicolumn{2}{c|}{Ped. (L1)} & \multicolumn{2}{c|}{Ped. (L2)} & 
\multicolumn{2}{c|}{Cyc. (L1)} & \multicolumn{2}{c}{Cyc. (L2)} \\
 & &mAP & mAPH & mAP & mAPH & mAP & mAPH & mAP & mAPH & mAP & mAPH & mAP & mAPH \\ 
\noalign{\smallskip}
\hline
\noalign{\smallskip}
MVF \cite{chen2017multi}& One & 62.93 & - & - & - & 65.33 & - & - & - & - & - & - \\
3D-MAN \cite{yang20213d} & Multi & 69.03 & 68.52 & 60.16 & 59.71 & 71.71 & 67.74 & 62.58 & 59.04 & - & - & - & -\\
RCD \cite{bewley2020range} & Two & 69.59 & 69.16 & - & - & - & - & - & - & - & - & - & - \\
$\dagger$SECOND \cite{yan2018second} & One & 72.27 & 71.69 & 63.85 & 63.33 & 68.70 & 58.18 & 60.72 & 51.31 & 60.62 & 59.28 & 58.34 & 57.05 \\
$\dagger$PointPillar \cite{lang2019pointpillars} & One & 56.62 & - & - & - & 59.25 & - & - & - & - & - & - & - \\
LiDAR R-CNN \cite{li2021lidar} & Two & 73.50 & 73.00 & 64.70 & 64.20 & 71.20 & 58.70 & 63.10 & 51.70 & 68.60 & 66.90 & 66.10 & 64.40 \\
RangeDet \cite{fan2021rangedet} & One & 72.85 & - & - & - & 75.94 & - & - & - & 65.80 & - & - & - \\
MVF++ \cite{qi2021offboard} & One & 74.64 & - & - & - & 78.01 & - & - & - & - & - & - & - \\
RSN \cite{sun2021rsn} & Two & 75.10 & 74.60 & 66.00 & 65.50 & 77.80 & 72.70 & 68.30 & 63.70 & - & - & - & - \\
Voxel R-CNN \cite{deng2020voxelrcnn} & Two & 75.59 & - & 66.59 & - & - & - & - & - & - & - & - & - \\
CenterPoint \cite{yin2021center} & Two & 76.70 & 76.20 & 68.80 & 68.30 & 79.00 & 72.90 & 71.00 & 65.30 & - & - & - & - \\
Part-A$^2$ \cite{shi2020points} & Two & 77.05 & 76.51 & 68.47 & 67.97 & 75.24 & 66.87 & 66.18 & 58.62 & 68.60 & 67.36 & 66.13 & 64.93 \\
PV-RCNN \cite{shi2020pv}& Two & 77.51 & 76.89 & 68.98 & 68.41 & 75.01 & 65.65 & 66.04 & 57.61 & 67.81 & 66.35 & 65.39 & 63.98 \\
AFDetV2 \cite{Hu2021AFDetV2RT} & One & 77.64 & 77.14 & 69.68 & 69.22 & 80.19 & \textbf{74.62} & 72.16 & \textbf{66.95} & \textbf{73.72} & \textbf{72.74} & \textbf{71.06} & \textbf{70.12} \\ \hline
PillarNet-vgg & One & 77.41 & 76.86 &69.46 & 68.96 & 78.30 & 70.32 & 70.00 & 62.62 & 69.48 & 68.35 & 66.87 & 65.78 \\
PillarNet-18 & One & 78.24 & 77.73 & 70.40 & 69.92 & 79.80 & 72.59 & 71.57 & 64.90 & 70.40 & 69.29 & 67.75 & 66.68 \\
PillarNet-34 & One & \textbf{79.09} & \textbf{78.59} & \textbf{70.92} & \textbf{70.46} & \textbf{80.59} & 74.01 & \textbf{72.28} & 66.17 & 72.29 & 71.21 & 69.72 & 68.67 \\ \hline
\multicolumn{14}{l}{Two-frame 3D detection results of PillarNet variants for reference.} \\
PillarNet-vgg & One & 78.26 & 77.73 & 70.56 & 70.07 & 80.88 & 77.53 & 72.73 & 69.58 & 67.72 & 66.88 & 65.54 & 64.72\\
PillarNet-18 & One & 79.59 & 79.06 & 71.56 & 71.08 & 82.11 & 78.82 & 74.49 & 71.35 & 70.41 & 69.57 & 68.27 & 67.46\\
PillarNet-34 & One & 79.98 & 79.47 & 72.00 & 71.53 & 82.52 & 79.33 & 75.00 & 71.95 & 70.51 & 69.69 & 68.38 & 67.58 \\ \hline    
\end{tabular}
}
\end{table}

\subsection{Overall Results}

{\flushleft \bf Evaluation on nuScenes \textit{test} set.}
We also compare our PillarNet variants with previous LiDAR-only non-ensemble methods on the nuScenes \textit{test} set. 
As shown in Table \ref{tab:nuscenes_test}, 
all our PillarNet-vgg/18/34 go beyond the stage-of-the-art methods by a large margin while running at a real-time speed of 14, 13 and 12 FPS, respectively.
In addition, the promising results of PillarNet variants validate the good scalability of our PillarNet, where the performance behaves more favorably as the computational complexity rises.
Typically, PillarNet-18 surprisingly surpasses the most advanced AFDetV2 by +2.3\% NDS or +2.6\% mAP.
To the best of our knowledge, PillarNet-vgg/18/34 surpasses all the published LiDAR-only non-ensemble methods on the nuScenes Detection leaderboard on Mar 7, 2022.
From this point on, PillarNet achieves new state-of-the-art performance using only 2D convolutions.

{\flushleft \bf Evaluation on Waymo Open Dataset \textit{test} set.}
We compare our PillarNet variants with previous methods on the Waymo Open Dataset \textit{test} set.
Table \ref{tab:waymo_test} contains two groups, where the upper group is single-frame LiDAR-only non-ensemble methods and the bottom group is multi-frame LiDAR-only non-ensemble methods.
Our PillarNet-34 outperforms all the previous single-frame and multi-frame LiDAR-only models for the vehicle and pedestrian categories while running at a speed of 19 FPS separately.
Our lightweight PillarNet-vgg still achieves the comparable performance for the vehicle while running at a faster speed of 24 FPS. 
Using merely 2D convolutions, our real-time PillarNet variants are suitable for onboard deployment.

{\flushleft \bf Evaluation on Waymo Open Dataset \textit{val} set.}
We compare our PillarNet variants with all published single-frame LiDAR-only non-ensemble methods on Waymo \textit{val} set in Table \ref{tab:waymo_val}. We also present the performance of PillarNet variants using two-frame-merged LiDAR points for reference.
Typically, PillarNet-18 achieves the state-of-the-art performance on the vehicle category, making it a viable replacement for previous state-of-the-art 3D voxel-based methods.
Our PillarNet-34 outperforms previous state-of-the-art works with remarkable performance gains (+1.24 for the vehicle in terms of mAPH of LEVEL\_2 difficulty).
Excluding the latest voxel-based detector AFDetV2 with self-calibrated module and channel-wise and spatial-wise attention, PillarNet-34 outperforms the previous one-stage and two-stage 3D detectors for the vehicle and pedestrian detection while operating at super real-time speed.
With the two-frame input, PillarNet on variant backbones consistently show the superior performance compared with their single-frame counterparts. However, for the cyclist detection, two-frame results are not the best. The reason may be the unbalanced sample distribution of three categories. 
The number of vehicles, pedestrians and cyclists scattered in the Waymo train set are 4352210, 2037627 and 49518 respectively.
The training process using two frames aggravates the adverse effect, and this  issue may be alleviated by addressing the unbalanced sample distribution.

\begin{table}[t]
\centering
\caption{The analysis of each component of PillarNet with the same training schedules as SECOND and also comparison with the two baselines (\textit{i.e.}, PointPillars and SECOND) on nuScenes \textit{val} dataset. $\dagger$: reported by used codebase.}
\label{tab:ab_compare}
\begin{tabular}{l|c|cc|ccccc} 
\hline\noalign{\smallskip}
Methods & FPS & mAP & NDS & mATE & mASE & mAOE & mAVE & mAAE  \\ \hline
$\dagger$CenterPoint-PointPillars \cite{lang2019pointpillars} & 31 & 50.26 & 60.22 & 31.32 & 25.94 & 39.50 & 32.54 & 19.79 \\
$\dagger$CenterPoint-SECOND \cite{yan2018second} & 8 & 59.56 & 66.76 & 29.22 & 25.51 & 30.24 & 25.91 & 19.34 \\
PointPillars (0.075m)$^{\textcolor{red}{1}}$ & 9 & 48.63 & 59.51 & 30.70 & 26.35 & 35.81 & 35.52 & 19.70 \\
\noalign{\smallskip}
\hline
\noalign{\smallskip}
PillarNet-18(neckv1) & 17 & 57.87 & 66.16 & 29.88 & 26.05 & 27.86 & 25.78 & 18.14 \\
PillarNet-18(neckv2) & 16 & 58.53 & 66.41 & 29.82 & 26.05 & 29.56 & 24.53 & 18.65 \\ 
PillarNet-18(neckv2-\textit{D}) & 16 & 59.40 & 66.96 & 29.07 & 25.86 & 29.61 & 24.69 & 18.17 \\
PillarNet-18(neckv3) & 16 & 59.48 & 67.15 & 29.03 & 25.83 & 27.54 & 24.54 & 18.90 \\
\hline
PillarNet-18(OD-IoU) & 16 & 59.51 & 67.09 & 28.51 & 25.57 & 29.31 & 24.63 & 18.64  \\
PillarNet-18(OD-GIoU) & 16 & 59.69 & 67.35 & 28.50 & 25.78 & 27.57 & 24.74 & 18.37 \\
PillarNet-18(OD-DIoU) & 16 & 59.72 & 67.39 & 28.40 & 25.81 & 27.38 & 24.67 & 18.41 \\
PillarNet-18(IoU) & 16 & 59.82 & 67.16 & 28.92 & 25.53 & 28.37 & 25.63 & 19.07 \\ \hline
PillarNet-18 & 16 & 59.90 & 67.39 & 27.72 & 25.20 & 28.93 & 24.67 & 19.11 \\
\hline
\end{tabular}
\end{table}

\subsection{Ablation Studies}

In this section, we investigate the individual components of the proposed PillarNet with extensive ablation experiments on the \textit{val} set of nuScenes Dataset.

{\flushleft \bf Analysis of PillarNet improvements.}
The key contribution can be summarized into two parts: the designed PillarNet architecture (\textit{i.e.}, encoder and neck networks) and the IoU-related modules (\textit{i.e.}, Orientation-Decoupled IoU (OD-IoU) regression loss and IoU-Aware rectification).
To analyze how our designed encoder and neck networks improve the 3D detection performance, we use the same hyper-parameters settings as CenterPoint-SECOND.

\textit{Encoder network.}
Compared with CenterPoint-PointPillars \cite{lang2019pointpillars} ($1^{st}$ row of Table \ref{tab:ab_compare}), our newly introduced encoder network can significantly improve the detection performance by about +7.61\% mAP and +5.94\% NDS.
Using the heavy encoder with extra stage 5 in $3^{rd}$ to $6^{th}$ rows can boost the 3D detection performance by a large margin.
Therein, the enriched semantic features from encoder stage 5 in $4^{th}$ to $5^{th}$ rows perform better than the aggressive fusion strategy in $6^{th}$ row.

\textit{Neck network.}
Compared with the naive neck module neckv1, as shown in Table \ref{tab:ab_compare}, our fusion design from $4^{th}$ to $6^{th}$ rows with a group of convolution layers can improve detection performance by a large margin.
%
The performance difference between neckv2 and neckv2-\textit{D} shows that the dense convolutions enable stronger semantic abstraction at the object center than its sparse counterparts, due to LiDAR points sparsely scattering on the surface of the objects. 

\textit{OD-IoU regression loss.}
All three types of losses (\textit{i.e.}, OD-IoU, OD-GIoU and OD-DIoU) play a role in the critical positioning accuracy, while the OD-DIoU loss brings a maximum boost with +0.24\% mAP or +0.24\% NDS. 

\textit{IoU-Aware rectification.} The IoU-Aware rectification  alleviates the misalignment between localization confidence and classification score. 
Adding IoU-Aware rectification benefits the IoU-based mAP with +0.34\% increase.

\setlength{\tabcolsep}{1.4pt}
\begin{table}[t]
\parbox{0.38\linewidth}{
\centering
\caption{The effect of different PillarNet variants by detaching two IoU-related modules.}
\label{tab:ab_bone}
\begin{tabular}{c|c|c|cc}
\hline
Models & FPS & mAP & NDS \\ \hline
PillarNet-vgg & 16 & 57.67 & 65.71 \\
PillarNet-18 & 16 & 59.41 & 67.09 \\
PillarNet-34 & 14 & 59.98 & 67.50 \\
\hline
\end{tabular}
}
\hfill
\parbox{0.6\linewidth}{
\centering
\caption{The effect of different pillar sizes and its associated stages in PillarNet-18 encoder.}
\label{tab:ab_pillar}
\begin{tabular}{c|c|c|cc}
\hline
Pillar size & FPS & encoder stages & mAP & NDS  \\ \hline
0.075m & 16 & (1x 2x 4x 8x 16x) & 59.48 & 67.15 \\
0.075*2m & 16 & (2x 4x 8x 16x) & 58.70 & 66.56 \\
0.075*4m & 16 &(4x 8x 16x) & 57.87 & 66.05 \\
0.075*8m & 14 & (8x 16x) & 55.37 & 64.20 \\
\hline
\end{tabular}
}
\end{table}

{\flushleft \bf Analysis of model variants.}
We investigate PillarNet-vgg/18/34 with different model complexity by detaching IoU-related modules for a clean comparison.
Table \ref{tab:ab_bone} shows that our PillarNet architecture can benefit from increasing model capacity with slightly more FLOPs and inference time. 
The good scalability can lead the pathway for deployment according to practical needs.

{\flushleft \bf Analysis of pillar sizes.}
We investigate pillar size on detection performance by detaching IoU-related modules for a clean comparison.
Specifically, we castrate the associated encoder stages to suit particular pillar scale, where a larger pillar size requires fewer encoder stages.
From Table \ref{tab:ab_pillar} and Fig \ref{fig:performance}, we can see that PillarNet 
 benefit more from finer pillar scale and deeper pillar feature encoding.
The much higher performance of PillarNet with 0.3m and 0.6m over PointPillars with 0.2m manifests the effectiveness of our architectural design. 
Moreover, PointPillars \cite{lang2019pointpillars} with 0.075m in $3^{rd}$ row of Table \ref{tab:ab_compare} performs slightly worse than that of 0.2m. That is because the used lightweight encoder network hinders the gain from small pillar size. This also implies the importance of hierarchical pillar feature encoding of PillarNet for better performance with limited resource costs.


{\flushleft \bf Runtime analysis.}
We analyze inference runtime by fairly comparing with two baseline counterparts.
our PillarNet-18 achieves a good speed-accuracy trade-off with 16 FPS than CenterPoint-SECOND of 8 FPS on nuScenes Dataset, and 21 FPS than CenterPoint-PointPillars of 19 FPS on Waymo Open Dataset.
%
The slow inference speed for PillarNet with coarser pillar size and reduced encoder stages in Table \ref{tab:ab_pillar} may be due to the fact that cuda $atomic \ max$ operation struggles to handle more inside points per pillar based on global memory.
This issue can be alleviated by the input point cloud sub-sampling or other efficient operation (\textit{e.g.,} streaming pillarization as \cite{chen2021polarstream}).

\section{Conclusions}

In this work, we propose a real-time and high-performance one-stage 3D object detector. From the perspective of ``encoder-neck-head" architecture design, PillarNet achieves the scalability and flexibility for the hard-balanced pillar size and model complexities. 
We expect that our findings will stimulate further research into pillar-based point cloud representation learning.

{\flushleft \bf Acknowledgements.}
This work was supported in part by NSFC (61906119), and Shanghai Municipal Science and Technology Major Project (2021SHZDZX0102). We thank Huawei Noah’s Ark Lab gratefully for sponsoring large-scale GPUs. 




\clearpage
%
%
\bibliographystyle{splncs04}
\bibliography{egbib}
\end{document}